# Parts-of-Speech Tagger Errors Do Not Necessarily Degrade Accuracy in Extracting Information from Biomedical Text

**Maurice HT Ling[1], Christophe Lefevre [1,2], Kevin R Nicholas[1]**

[1]CRC for Innovative Dairy Products, Department of Zoology, The University of Melbourne, Australia
[2]Victorian Bioinformatics Consortium, Monash University, Australia

**Corresponding email: mauriceling@acm.org**

## Abstract

**Background:** An ongoing assessment of the literature is difficult with the rapidly increasing volume of research publications and limited effective information extraction tools which identify entity relationships from text. A recent study reported development of Muscorian, a generic text processing tool for extracting protein-protein interactions from text that achieved comparable performance to biomedical-specific text processing tools. This result was unexpected since potential errors from a series of text analysis processes is likely to adversely affect the outcome of the entire process. Most biomedical entity relationship extraction tools have used biomedical-specific parts-of-speech (POS) tagger as errors in POS tagging and are likely to affect subsequent semantic analysis of the text, such as shallow parsing. This study aims to evaluate the parts-of-speech (POS) tagging accuracy and attempts to explore whether a comparable performance is obtained when a generic POS tagger, MontyTagger, was used in place of MedPost, a tagger trained in biomedical text. **Results:** Our results demonstrated that MontyTagger, Muscorian's POS tagger, has a POS tagging accuracy of 83.1% when tested on biomedical text. Replacing MontyTagger with MedPost did not result in a significant improvement in entity relationship extraction from text; precision of 55.6% from MontyTagger versus 56.8% from MedPost on directional relationships and 86.1% from MontyTagger compared to 81.8% from MedPost on nondirectional relationships. This is unexpected as the potential for poor POS tagging by MontyTagger is likely to affect the outcome of the information extraction. An analysis of POS tagging errors demonstrated that 78.5% of tagging errors are being compensated by shallow parsing. Thus, despite 83.1% tagging accuracy, MontyTagger has a functional tagging accuracy of 94.6%. **Conclusions:** The POS tagging error does not adversely affect the information extraction task if the errors were resolved in shallow parsing through alternative POS tag use.

## 1. Introduction

PubMed currently indexes more than 17.5 million papers that includes 1 million papers added in both 2006 and the first half of 2007. This trend of increased volume of research papers makes it difficult for researchers to maintain a productive assessment of relevant literature. Information extraction (IE) has been used as a tool to analyze biological text to derive assertions, such as entity interactions (Abulaish and Dey, 2007). To date, there has been a number of IE tools to extract entity interactions from published text, such as MedScan (Novichkova et al., 2003), Arizona Relation Parser (Daniel et al., 2004), BioRAT (David et al., 2004) and Santos et al. (2005).

A recent article by Ling et al. (2007) has classified entity interaction IE tools by whether tools are developed with biological text in mind or adapted generic tools for biological text. Ling et al. (2007) developed Muscorian, a tool to extract protein-protein interactions from text. They also demonstrated that a generic text analysis tool chain, MontyLingua (Liu and Singh, 2004; Ling, 2006), incorporated into a two-layered generic-specialized architecture as explained in MedScan (Novichkova et al. 2003), can give rise to comparable performance in entity interaction extraction compared to those IE systems that modified existing systems, such as BioRAT (David et al., 2004), Chilibot (Chen and Sharp, 2004) and Santos et al. (2005). One of the common features of both classes of tools defined by Ling et al. (2007) is the specialization of the part-of-speech (POS) tagger. For example, Arizona Relation Parser (Daniel et al., 2004) re-trained Brill tagger (Brill, 1995) and Chilibot (Chen and Sharp, 2004) re-trained TnT tagger (Brants, 2000). POS tagging is a process of assigning grammatical roles of each word and punctuation in the source sentence. This plays a critical role in subsequent text processing tasks, such as shallow parsing, where the sequence of POS tags were used instead of the original sequence of words. At the same time, it was known that errors in POS tagging often results in misunderstanding of the sentence (Kodratoff et al., 2005; Amrani et al., 2005).

Muscorian (Ling et al., 2007) makes use of a generic POS tagger as part of MontyLingua (Ling, 2006; Liu and Singh, 2004) and performs at a comparable level to IE tools using POS taggers trained on biomedical text. This contradicts the common view that "*error propagation through cascades of processors may in aggregate severely degrade performance on the final task*" as stated in the *Call for Papers for the Tenth Conference on Natural Language Processing 2006 (CoNLL-X)*. Tateisi and Tsujii (2004) have demonstrated that generic POS taggers are only about 83% accurate when used to tag biomedical text. This suggests that MontyTagger, the generic POS tagger in MontyLingua, is unlikely to perform as well as taggers trained on biomedical text, such as MedPost (Smith et al., 2004). Therefore, it is likely that the above mentioned contradiction is resolved at the step immediately downstream to POS tagging, the shallow parsing. In MontyLingua shallow parsing (Ling et al., 2007), the input sentence is broken into noun phrase and verb phrase. The process of shallow parsing can be seen as a collapse of a sequence of POS tags into 2 groups; hence, we expect high level of permissible substitution of POS tags within related classes. We term this permissible substitution as "alternate POS tag use".

This study compares the performance of MedPost (Smith et al., 2004) with the generic POS tagger, MontyTagger (Liu and Singh, 2004), in Muscorian (Ling et al., 2007) and illustrates a case whereby POS tagging error does not adversely affect the final information extraction task if the errors were resolved in shallow parsing through alternate POS tag use.

## 2. Methods
### 2.1. Evaluating POS Tagging and Information Extraction Performance
MontyTagger was evaluated on its own using MedPost corpus (Smith et al., 2004) and its accuracy as the percentage of the number of correctly tagged tokens (words and punctuations) in the total number of tokens (n=182399). MedPost tagger was swapped in place of MontyTagger by modifying MontyLingua's *jist()* and *jist_predicates()*

functions to *mpjist()* and *mpjist_predicates()*, giving MedPost-MontyLingua Muscorian:

```
def jist(self,text):                              def jist_predicates(self,text):
    sentences = self.split_sentences(text)            infos = self.jist(text)
    tokenized = map(self.tokenize,sentences)          svoos_list = []
    tagged = map(self.tag_tokenized,tokenized)        for info in infos:
    chunked = map(self.chunk_tagged,tagged)               svoos =
    extracted = map(self.extract_info,chunked)      info['verb_arg_structures_concise']
    return extracted                                      svoos_list.append(svoos)
                                                      return svoos_list
```

to

```
def mpjist(self,text):                                mpout = open('temp' + os.sep +
    sentences = self.split_sentences(text)                str(outfilename), 'r')
    tokenized = map(self.tokenize,sentences)          tagged = mpout.readlines()
    sourcefilename =                                  mpout.close()
        random.random()*1000000000                    chunked = map(self.chunk_tagged,tagged)
    outfilename =                                     extracted = map(self.extract_info,chunked)
        random.random()*1000000000000                 return extracted
    source = open('temp' + os.sep +
        str(sourcefilename), 'w')                 def mpjist_predicates(self,text):
    source.writelines(tokenized)                      infos = self.mpjist(text)
    source.close()                                    svoos_list = []
    os.popen(os.getcwd() + os.sep +                   for info in infos:
        'medpost/medpost -text -token -penn <             svoos =
        temp' + os.sep + str(sourcefilename) + '>       info['verb_arg_structures_concise']
        temp' + os.sep + str(outfilename))                svoos_list.append(svoos)
                                                      return svoos_list
```

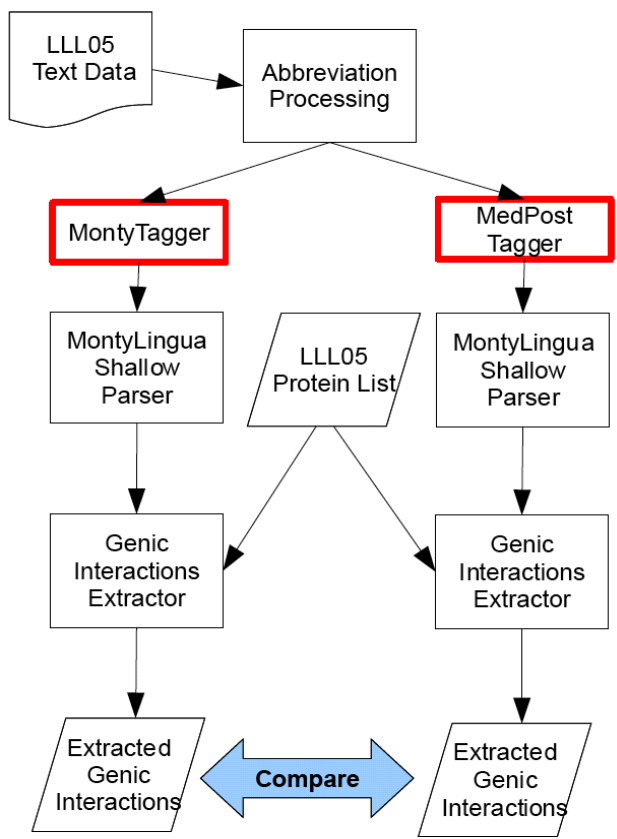

Figure 1. <u>Flowchart of evaluation procedure for Muscorian with native MontyLingua and MedPost-MontyLingua.</u> LLL05 test data was processed for abbreviations before feeding into each system and the extracted genic interactions (output) were evaluated for precision and recall.

MedPost-MontyLingua Muscorian's IE performance was evaluated using Learning Languages in Logic 2005 test data (Cussens and Nedellec, 2005) in the same manner as Muscorian (Ling et al., 2007) and the performances were compared (Figure 1).

## 2.2. Analysis of POS Tagging Errors

Wrongly tagged tokens from MontyTagger's output were first grouped by their original tags in MedPost corpus (Smith et al., 2004), then sub-grouped by MontyTagger's assigned tags (the wrong tag) and arranged in decreasing order based on the numbers of tags in both main and sub-group. First 80% of the tags in the main group where first 90% of the wrongly assigned tags were chosen for further error analysis. Each of the pairs of original tag and wrongly assigned tag were analysed with respect to the regular expressions in MontyREChunker (Ling et al., 2007), the shallow parser in MontyLingua, for the effects of the wrongly assigned tags on the operations of the shallow parser.

## 3  Results

### 3.1  Evaluating POS Tagging and Information Extraction Performance

Evaluating MontyTagger on MedTag corpus demonstrated correct tagging in 151663 of the tags representing 83.1% tagging accuracy. Using the LLL05 evaluation corpus, Muscorian with MedPost-MontyLingua on directional relationship was found to be 56.8% precise with 24.8% recall, while nondirectional relationship was estimated to be 81.8% precise with 35.6% recall (Table 1).

|  | *Directional Relationships* |  | *Nondirectional Relationships* |  |
| --- | --- | --- | --- | --- |
|  | **MontyLingua** | **mpMontyLingua** | **MontyLingua** | **mpMontyLingua** |
| Precision | 55.6% | 56.8% | 86.1% | 81.8% |
| Recall | 19.8% | 24.8% | 30.7% | 35.6% |
| F-Score | 0.292 | 0.345 | 0.453 | 0.496 |

Table 1. Summary of Muscorian's performances evaluated using Learning Languages in Logic 2005 data (Cussens, 2005).

### 3.2.  Analysis of POS Tagging Errors

Comparison of the reference tags (MedPost corpus) with the wrongly assigned tags from MontyTagger showed the 30736 wrongly assigned tags (52.3%, n=16067) should be tagged as nouns (tag: 'NN'), 15.8% (n=4865) should be tagged as 'JJ' (adjectives), and the next four most common wrongly assigned tags were 'NNS' (n=1987, 6.5%), 'SYM' (n=1496, 4.9%), 'VBP' (n=1470, 4.8%), and 'VBD' (n=745, 2.4%). These six reference tags (NN, JJ, NNS, SYM, VBP, VBN) accounted for 26630 (86.6%) of the wrongly assigned tags, while the rest of the errors (n=4106) were distributed across 25 tags. Six tags (TO, :, (, ), WP, ,) were correctly assigned in every instance in this evaluation. A tabulation of errors is shown in Table 2 and a table providing the definition of each POS tag is given in Table 3. The confusion matrix can be found at http://ib-dwb.sf.net/Muscorian/MedPost_confuse.txt.

| Tag | % Corpus | % Error in Total Error | % Error in Tag | Tag | % Corpus | % Error in Total Error | % Error in Tag |
|---|---|---|---|---|---|---|---|
| NN | 28.56 | 52.27 | 30.84 | VBG | 0.64 | 0.06 | 1.59 |
| IN | 13.49 | 1.08 | 1.33 | : | 0.54 | 0.00 | 0.00 |
| JJ | 10.47 | 15.81 | 25.44 | MD | 0.43 | 0.01 | 0.2 |
| DT | 7.77 | 0.56 | 1.16 | WDT | 0.45 | 0.19 | 6.70 |
| NNS | 7.75 | 6.45 | 14.03 | , | 0.39 | 0.00 | 0.00 |
| CC | 6.66 | 1.30 | 3.29 | PRP$ | 0.28 | 0.01 | 0.40 |
| . | 3.67 | 0.01 | 0.03 | FW | 0.26 | 0.96 | 61.39 |
| CD | 3.13 | 2.02 | 10.84 | WRB | 0.23 | 0.59 | 43.33 |
| VBN | 3.05 | 1.70 | 10.13 | JJR | 0.17 | 0.17 | 17.74 |
| VBD | 2.81 | 2.42 | 14.56 | NNP | 0.14 | 0.03 | 3.53 |
| RB | 2.57 | 1.72 | 9.49 | EX | 0.08 | 0.01 | 1.38 |
| ) | 1.89 | 0.00 | 0.00 | POS | 0.06 | 0.06 | 15.31 |
| ( | 1.88 | 0.00 | 0.00 | WP | 0.06 | 0.00 | 0.00 |
| VBP | 1.98 | 4.78 | 41.26 | JJS | 0.05 | 0.02 | 6.60 |
| TO | 1.55 | 0.00 | 0.00 | RBS | 0.05 | 0.01 | 4.40 |
| VBZ | 1.54 | 0.45 | 5.20 | " | 0.03 | 0.19 | 100.00 |
| SYM | 1.07 | 4.87 | 76.43 | `` | 0.03 | 0.19 | 100.00 |
| PRP | 0.88 | 1.61 | 30.59 | PDT | 0.02 | 0.11 | 100.00 |
| VB | 0.74 | 0.05 | 1.11 | RBR | 0.01 | 0.03 | 44.44 |

Table 2. <u>Percentage breakdown of POS tags in MedTag corpus and errors in MontyTagger as percentage of POS tags assignation.</u> This table tabulates the POS tagging errors made by MontyTagger on MedTag corpus and the order is according to the abundance of each tag in the MedTag corpus. For example, 'NN' is the most abundant tag accounting for 28.56% or 52093 of MedTag corpus of 182399 tokens. Of which, 3084% (16067 of 52093) of the 'NN' tokens in MedTag corpus were wrongly assigned to a different POS tag by MontyTagger which accounted for 52.27% of the total wrongly assigned POS tag of 30736 tokens.

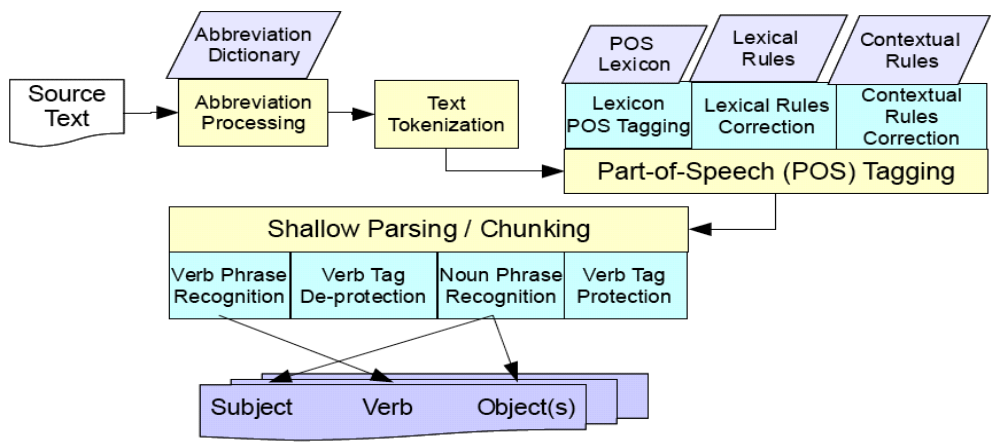

Figure 2. Muscorian's generalization layer, from source text to subject-verb-object(s) structures (Ling et al., 2007).

| Tag | Description | Tag | Description |
| --- | --- | --- | --- |
| CC | Coordinating conjunction | PRP$ | Possessive pronoun |
| CD | Cardinal number | RB | Adverb |
| DT | Determinant | RBR | Adverb, comparative |
| EX | Existential *there* | RBS | Adverb, superlative |
| FW | Foreign word | RP | Particle |
| IN | Preposition or subordinating conjunction | SYM | Symbol |
| JJ | Adjective | TO | to |
| JJR | Adjective, comparative | UH | Interjection |
| JJS | Adjective, superlative | VB | Verb, base form |
| LS | List item marker | VBD | Verb, past tense |
| MD | Modal | VBN | Verb, past participle |
| NN | Noun, singular or mass | VBG | Verb, gerund or present participle |
| NNS | Noun, plural | VBP | Verb, non-3$^{rd}$ person singular present |
| NNP | Proper noun, singular | VBZ | Verb, 3$^{rd}$ person singular present |
| NNPS | Proper noun, plural | WDT | Wh-determiner |
| PDT | Predeterminer | WP | Wh-pronoun |
| POS | Possessive ending | WP$ | Possessive wh-pronoun |
| PRP | Personal pronoun | WRB | Wh-adverb |

Table 3. Penn Treebank Tag Set without Punctuation Tags (Adapted from (Marcus et al., 1993))

An understanding of the general scheme of operations of MontyLingua as described in Ling et al. (2007), especially downstream process of POS tagging, the process of shallow parsing by MontyREChunker (MontyLingua's shallow parser) is crucial in our error analysis (Figure 2). Source text (abstracts) were processed for abbreviations and tokenized into sentences, then words and punctuations, before POS tagging. In POS tagging, each token was tagged first using a lexicon and corrected using lexical and contextual rules. This was where the output was 83.1% accurate compared to 96.9% in MedPost. POS tagging could be seen as a reduction of potentially unlimited human English words into 45 "words" or tags using knowledge of English grammar, and the sequence of tags was the input to the shallow parser, MontyREChunker. Firstly, verb tags (VBD, VBG and VBN) were protected by suffixing the tags to prevent interference in subsequent noun phase recognition (Ling et al., 2007). This meant that wrong tagging between these three tags, such as VBD was erroneously tagged as VBN, had no effect on this process. However, wrong tagging of any of these three tags to any of the other 42 tags or the other way around will be detrimental to this process. Secondly, noun phrases were recognized by the following regular expression (according to Python regex engine in the Python standard library):

((((PDT )?(DT |PRP[$] |WDT |WP[$] )(VBG |VBD |VBN |JJ |JJR |JJS |, |CC |NN |NNS |NNP |NNPS |CD )*(NN |NNS |NNP |NNPS |CD )+)|((PDT )?(JJ |JJR |JJS |, |CC |NN |NNS |NNP |NNPS |CD )*(NN |NNS |NNP |NNPS |CD )+)|EX |PRP |WP |WDT ) POS )?(((PDT )?(DT |PRP[$] |WDT |WP[$] )(VBG |VBD |VBN |JJ |JJR |JJS |, |CC |NN |NNS |NNP |NNPS |CD )*(NN |NNS |NNP |NNPS |CD )+)|((PDT )?(JJ |JJR |JJS |, |CC |NN |NNS |NNP |NNPS |CD )*(NN |NNS |NNP |NNPS |CD )+)|EX |PRP |WP |WDT )

A number of relationships that potentially contribute to reduced POS tagging errors were considered. Firstly, these four tags; DT, PRP[$], WDT, and WP[$]; were alternatives to each other and erroneous tagging between them had no effect on shallow parsing. Secondly, the ten tags; JJ, JJR, JJS, ",",CC, NN, NNS, NNP, NNPS, and CD; were alternatives to each other. Lastly, these four tags; EX, PRP, WP, and WDT; were alternatives to each other.

Subsequently, verb phases were de-protected by removing the suffix appended during the tag protection phase (Ling et al., 2007), followed by verb phrase recognition. This meant that verb tag protection had the highest precedence, followed by noun phrase recognition, and then verb phrase recognition. This meant that nullified errors in higher precedence would not affect downstream processes. Verb phrases were recognized by the following regular expression:

(RB |RBR |RBS |WRB )*(MD )?(RB |RBR |RBS |WRB )*(VB |VBD |VBG |VBN |VBP |VBZ )(VB |VBD |VBG |VBN |VBP |VBZ |RB |RBR |RBS |WRB )*(RP )?(TO (RB )*(VB |VBN )(RP )?)?

In terms of compensation for POS tagging errors, this meant that the four tags; RB, RBR, RBS, and WRB; and these six tags; VB, VBD, VBG, VBN, VBP, and VBZ; were alternatives to each other. However, verb phrase required terminal VB or VBN, which meant that although verb tag protection allowed for interchangeable use of VBN, VBG and VBD, erroneous tagging of VBG and VBD to VBN or VBN to VBG or VBD would be detrimental to verb phrase recognition.

A deeper analysis was undertaken to examine the errors in each reference tag (tabulated in Table 4). Firstly, by grouping close POS types, for example 'NN', 'NNP', and 'NNS' were all nouns, wrong sub-type assignation, such as 'NN' assigned as 'NNP' and 'NNS' assigned as 'NNP', accounted for 55% of the errors (n=14634). Secondly, 58% (n=2818) of 'JJ' (adjective) errors were resulted by tagging as noun (NN and NNP) while 34.9% (n=1698) of the 'JJ' errors were tagged as verb (VBN and VBG). Thirdly, about 5% (n=941) of 'NN' (noun) errors were tagged as cardinal numbers (CD). Fourthly, plural nouns accounted for 51.6% (n=1026) of 'NNS' (singular noun) errors. Fifthly, 48.7% (n=729) and 39.3% (n=587) of 'SYM' (symbol) errors were either not assigned or assigned as 'NN' (noun) respectively. Lastly, 87% (n=1927) of verb errors (VBP and VBD) were due resolution of tenses, such as non-third party singular present tense (VBP) was assigned as infinite verb form (VB).

Error breakdown (in Table 4) demonstrated erroneous POS tagging by MontyTagger in 31 tags, with 6 tags having no errors. A total of 6 of the 32 tags (19.4%) accounted for 86.6% (n=26630) of the total errors and were chosen for further analysis. Applying these error nullification rules to each of the examined erroneous tags (86.6% of the errors), it was found that 78.6% of the errors had no effect on shallow parsing. A tabulated analysis is shown in Table 4.

| *Reference Tag* | *Wrongly Assigned Tag* | *Number of Wrong Assignation* | *Cummulative Frequency for Reference Tag* | *Impact on Shallow Parsing?* |
|---|---|---|---|---|
| NN (16067) | NNP | 10865 | 67.6% | No, NNP was an alternative match to NN in noun phrase recognition |
| | JJ | 2527 | 83.4% | No, JJ was an alternative match to NN in noun phrase recognition |
| | CD | 941 | 89.2% | No, CD was an alternative match to NN in noun phrase recognition |
| | VBG | 812 | 94.3% | Yes, protected verb tag |
| JJ (4865) | NN | 1600 | 32.9% | No, NN was an alternative match to JJ in noun phrase recognition |
| | NNP | 1218 | 58.0% | No, NNP was an alternative match to JJ in noun phrase recognition |
| | VBN | 1170 | 82.0% | Yes, protected verb tag |
| | VBG | 528 | 92.8% | Yes, protected verb tag |
| NNS (1987) | NNP | 1026 | 51.6% | No, NNP was an alternative match to NNS in noun phrase recognition |
| | NN | 701 | 86.9% | No, NN was an alternative match to NNS in noun phrase recognition |
| | VBZ | 128 | 93.4% | No, VBZ was an alternative |

| Reference Tag | Wrongly Assigned Tag | Number of Wrong Assignation | Cummulative Frequency for Reference Tag | Impact on Shallow Parsing? |
|---|---|---|---|---|
| | | | | match to NNS in noun phrase and was not a protected verb tag |
| SYM (1496) | Not Assigned | 729 | 48.7% | No, tokens not tagged were non-existent and SYM was not used in shallow parsing |
| | NN | 587 | 88.0% | Yes, NN was matched in noun phrase |
| | - | 115 | 95.7% | No, both tags was not used in shallow parsing |
| VBP (1470) | VB | 1249 | 85.0% | Yes, mandatory requirement of VB in verb phrase |
| | NN | 178 | 97.1% | No, NN was an alternative match to VBP in noun phrase |
| VBD (745) | VBN | 678 | 91.0% | Yes, mandatory requirement of VBN in verb phrase |
| | JJ | 34 | 95.6% | Yes, protected verb tag |

Table 4. <u>Error breakdown and analysis on the effects of six most commonly mis-assigned POS tags.</u> Six reference tags; NN, JJ, NNS, SYM, VBP, and VBD; which accounted for 86.6% of all wrong POS assignation by MontyTagger were chosen and in each tag, the assigned tags which accounted for 90% of the errors were chosen for further analysis. For example, of 16067 tags that were tagged as 'NN' in MedTag corpus, MontyTagger wrongly tagged 10865 tokens as 'NNP' and has no effect on shallow parsing, 2527 tokens as 'JJ' and has no effect on shallow parsing, 941 tokens as 'CD' and has no effect on shallow parsing, and 812 tokens as 'VBG' with an effect on shallow parsing. These 4 wrong tagging accounted for 94.3% of all 'NN' tag errors. This also meant that 922 'NN' tag errors (5.7%) were not further analyzed. A complete confusion matrix is given in http://ib-dwb.sf.net/Muscorian/MedPost-confuse.txt.

### 4.  Discussion

The precision and recall of native MontyLingua Muscorian for extracting genic interactions from the LLL05 data set (Cussens and Nedellec, 2005) was 55.6% and 19.7% (F-score = 0.29) respectively for directional interactions which is about 5% higher in precision and similar in recall to that reported in LLL05 (Cussens and Nedellec, 2005). The precision and recall was 86.1% and 30.7% (F-score = 0.45) respectively for nondirectional interaction. The term "directional" means that the direction of protein activity is non-commutative, for example, "proteinA activates proteinB" does not the same as "proteinB activates proteinA". However, nondirectional means that the protein activity is commutative, for example, "proteinA binds to proteinB" has no different biological significance than "proteinB binds to proteinA". This formed the baseline to evaluate a biomedical-specialized part-of-speech (POS) tagger (Smith et al., 2004) modification of Muscorian, MedPost-MontyLingua Muscorian. The main reason for examining this specialized POS tagger was that it was developed for biomedical information extraction systems (Daniel et al., 2004; Chen and Sharp, 2004) and POS tagging errors were known to be detrimental in understanding human text (Kodratoff et al., 2005; Amrani et al., 2005). In addition, POS tagger modification had been done in a number of biomedical information extraction systems, such as Jang et al. (2006) and Chilibot (Chen and Sharp, 2004).

Examining MontyLingua's source codes, the main function that processes text is the *jist_predicate()* function, which calls the *jist()* function to process text (tokenization, POS tagging and shallow parsing) and then to extract the resulting set of subject-verb-objects (SVO) from *jist*'s output (Ling et al., 2007). The Python codes for these two functions were as follows:

```
def jist(self,text):                                def jist_predicates(self,text):
    sentences = self.split_sentences(text)              infos = self.jist(text)
    tokenized = map(self.tokenize,sentences)            svoos_list = []
    tagged = map(self.tag_tokenized,tokenized)          for info in infos:
    chunked = map(self.chunk_tagged,tagged)                 svoos =
    extracted = map(self.extract_info,chunked)      info['verb_arg_structures_concise']
    return extracted                                        svoos_list.append(svoos)
                                                        return svoos_list
```

As observed, *jist()* function calls *tokenize* function to tokenize the text, *tag_tokenized* function to perform POS tagging, *chunk_tagged* function to perform shallow parsing, and finally, *extract_info* function to extract SVOs from the parsed text. The systematic structure of MontyLingua's codes, especially the *jist()* function had simplified the substitution of MontyTagger (by *tag_tokenized* function) with MedPost. This implied that any of the other components in the text analysis process, like shallow parser (by *chunk_tagged* function) could be easily exchanged.

The precision and recall of MedPost-MontyLingua Muscorian evaluated using the LLL05 data set (Cussens and Nedellec, 2005) were 56.8% and 24.8% (F-score = 0.35) respectively for directional interactions, and 81.8% and 35.6% (F-score = 0.50) respectively for nondirectional interaction. Our results showed that using MedPost in place of MontyLingua's POS tagger, MontyTagger, had improved the F-score by about 5% in both directional and nondirectional interactions extraction, and recall (24.8% versus 19.7% and 35.6% versus 30.7%). However, as reasoned in Ling et al. (2007), precision was more important than recall when extracted protein-protein interactions were used to support other biological analyses and the problem with mediocre recall is resolved with large volumes of text.

Our results indicated that MedPost-MontyLingua Muscorian outperformed un-modified-MontyLingua Muscorian in extracting directional genic interactions in terms of both precision and recall, suggesting that MedPost-MontyLingua Muscorian was more suited for this purpose. However, the precision of MedPost-MontyLingua Muscorian underperformed in extracting non-directional genic interactions, despite better recall. This suggested that errors in MontyTagger (un-modified-MontyLingua's POS tagger) resulted in more directional errors than that of MedPost. Given that our interest was in nondirectional interactions and precision was more important than recall in our case, un-modified-MontyLingua Muscorian was chosen for future work.

We conclude that our experimental results indicated that un-modified-MontyLingua Muscorian performed as well as MedPost-MontyLingua Muscorian for the purpose of processing biomedical text for the extraction of genic interactions. Thus, in contrary to the general assumption that generic text processing systems must be modified before being suitable for processing biological text for extracting genic interactions as evident from numerous systems to date, we presented a case study where comparable performance could be achieved by using generic text processing tools. This outcome is consistent with a previous study using un-modified MontyLingua for processing peer-reviewed economics papers (van Eck, 2005; van Eck and van Den Berg, 2005).

An initial evaluation of MontyTagger on MedTag Corpus (Smith et al., 2004) indicated 83.1% accuracy, which was considerably less than from MedPost's reported accuracy of 96.9% (Smith et al., 2004) and was close to the 83.0% tagging accuracy of a generic POS tagger on biomedical text (Tateisi and Tsujii, 2004). This result was expected as MontyTagger was not developed for biomedical text (Ling et al., 2007).

The POS tagging errors were expected to impact on performance of the entire text processing pipeline but this was not observed in our results. Instead, the precision of un-modified-MontyLingua Muscorian was comparable to that of MedPost-MontyLingua Muscorian on directional genic interactions (55.6% versus 56.8%) and un-modified-MontyLingua Muscorian outperformed MedPost-MontyLingua Muscorian on nondirectional genic interactions (86.1% versus 81.8%). Taken collectively the precision of both system and their respective POS tagging accuracies, seemed contradictory to general expectations as stated in the *Call for Papers for the Tenth Conference on Natural Language Processing 2006 (CoNLL-X)*.

An error analysis on MontyTagger was carried out in attempt to provide insight into resolving this contradiction. A likely hypothesis to explain why POS tagging errors did not derail the entire text processing pipeline was that the errors were nullified post-tagging. Text processing is used in Muscorian as a means to convert unstructured text into structured form for data mining - an extremely limited use of natural language processing compared to more complex uses, such as automated translation. As mentioned previously, POS tagging can be seen as a process of mapping potentially infinite number of words in the English language into a finite set of tags, based on their syntactic meanings. Shallow parsing, also known as chunking, can then be seen as a process which examines the sequence of tags and splits them into semantic phrases, of which verb phrase and noun phrase are of interest in this case. Given that MontyLingua's shallow parser parses the sequence of tags into 3 types of phrases (verb, noun, and adjectives), it is conceivable that a number of POS errors have no effect on shallow parsing.

Of the 182399 token in MedTag Corpus (Smith et al., 2004), 30736 were erroneously tagged by MontyTagger (16.9% error) spreading over 40 tags. The top 6 most common tag errors accounted for 86.6% of the total errors and were chosen for further evaluation. In each of the 6 most abundant error tags, the top 95% of the errors were examined.

The effects of each type of errors, such as 'NN' wrongly tagged to 'NNP', were examined by analyzing the routines for shallow parsing which uses Regular Expressions. It was found that in 26630 of the examined POS tagging errors, 20928 (78.6% of 26630) had no effect on the chunking process and the remaining 5703 errors adversely affected shallow parsing, which might account for lower recall of un-modified-MontyLingua Muscorian as compared to MedPost-MontyLigua Muscorian.

Therefore, despite a low POS tagging accuracy of 83.1% by MontyTagger, more than three-quarters of the errors had no detrimental effect on chunking, suggesting a "functional POS tagging accuracy" of at least 94.6%, which was relatively close to MedPost's reported 97% accuracy (Smith et al., 2004). This apparent high "functional

POS tagging performance" despite poor actual tagging accuracy might be the reason to explain un-modified-MontyLingua Muscorian's good performance in LLL05 test (Cussens and Nedellec, 2005) despite poor tagging accuracy compared to MedPost-MontyLigua Muscorian. This suggested that the nature of POS tagging errors might be more important than a single measure of POS tagging accuracy in a specific use of generic text processing tools where a shallow parser is involved. Therefore, it can be inferred that applications of biomedical literature analysis where a shallow parser is likely to be involved, such as extracting entity interactions and protein or molecule localization, POS tagging errors may not result in a decline in system performance.

At the same time, it is known that building domain-specific text processing tools requires much manual efforts (Jensen et al., 2006) suggesting that the cost and effort needed to train taggers specifically for biomedical text may not be needed, depending on the target application. However, it should also be cautioned that other applications or systems that do not involve shallow parser, such as Arizona Relation Parser (Daniel et al., 2004) which uses full sentence parsing, are likely to benefit from superior POS tagging accuracy of MedPost (Smith et al., 2004) and may experience degraded results from tagging errors.

MedTag Corpus (Smith et al., 2004) was used as a standard for evaluating MontyTagger. However, only 38 of the 45 tags in Penn Treebank Tag Set wereused to annotate the corpus while the tagged output of MontyTagger illustrated the use of 45 tags. This might suggest inconsistencies or errors in MedPost Corpus, which were found in other POS tagged corpora (Peshkin and Savova, 2003; Ratnaparkhi, 1996).

## 5. Conclusions

In summary, analysis of the effects of MontyTagger's errors on downstream shallow parsing by MontyREChunker illustrated that 78.6% of the examined errors had no effect on shallow parsing. This implied that although the POS tagging accuracy of MontyTagger on MedPost Corpus was 83.1%, a majority of the errors had no downstream effect; thus, the functional POS tagging accuracy of MontyTagger was between 94.6% and 96.9%. A good functional POS tagging accuracy despite poor POS tagging accuracy, with respect to shallow parsing, is a likely reason for a comparative performance in extracting protein-protein interactions from text using a domain-specific or a generic POS tagger.